%% file: root.tex
\pdfoutput=1
\documentclass[letterpaper, 10 pt, journal, twoside]{IEEEtran}  

\IEEEoverridecommandlockouts                              

\usepackage{graphicx} 
\usepackage{tabularx}
\usepackage{multirow}
\usepackage{hyperref}
\usepackage{xcolor}
\usepackage[flushleft]{threeparttable}

\title{HomeEmergency - Using Audio to Find and Respond to Emergencies in the Home
}

\author{James F. Mullen Jr,$^{1,2}$ Dhruva Kumar,$^{1}$ Xuewei Qi$^{1}$, \\ Rajasimman Madhivanan$^{1}$, Arnie Sen$^{1}$, Dinesh Manocha$^{2}$, and Richard Kim$^{1}$
\thanks{This material is based in part upon work supported by the National Science Foundation Graduate Research Fellowship Program under Grant No. DGE 2236417. Any opinions, findings, and conclusions or recommendations expressed in this material are those of the author(s) and do not necessarily reflect the views of the National Science Foundation.}
\thanks{$^{1}$Amazon Lab126
        {\tt\small dhruvkm@amazon.com, qixuewei@amazon.com, rajasimm@amazon.com,
        senarnie@amazon.com,
        richk@amazon.com}}%
\thanks{$^{2}$University of Maryland
        {\tt\small mullenj@umd.edu, dmanocha@umd.edu}}%
}

\begin{document}
\bstctlcite{IEEEexample:BSTcontrol}

\maketitle

\begin{abstract} In the United States alone accidental home deaths exceed 128,000 per year. 
Our work aims to enable home robots who respond to emergency scenarios in the home, preventing injuries and deaths. 
We introduce a new dataset of household emergencies based in the ThreeDWorld simulator. 
Each scenario in our dataset begins with an instantaneous or periodic sound which may or may not be an emergency. The agent must navigate the multi-room home scene using prior observations, alongside audio signals and images from the simulator, to determine if there is an emergency or not.

In addition to our new dataset, we present a modular approach for localizing and identifying potential home emergencies. Underpinning our approach is a novel probabilistic dynamic scene graph (P-DSG), where our key insight is that graph nodes corresponding to agents can be represented with a probabilistic edge. This edge, when refined using Bayesian inference, enables efficient and effective localization of agents in the scene. We also utilize multi-modal vision-language models (VLMs) as a component in our approach, determining object traits (e.g. flammability) and identifying emergencies. We present a demonstration of our method completing a real-world version of our task on a consumer robot, showing the transferability of both our task and our method. Our dataset will be released to the public upon this papers publication.

\end{abstract}


\input{Sections/1-Intro}
\input{Sections/2-Related}
\input{Sections/3-Dataset}
\input{Sections/4-Method}
\input{Sections/5-Experiments}
\input{Sections/6-Conclusion}


{\small
\bibliographystyle{IEEEtran}
\bibliography{refs}
}

\end{document}

%% file: Sections/1-Intro.tex
\section{Introduction}

 \begin{figure}[t]
     \centering
     \includegraphics[width= 0.7\linewidth]{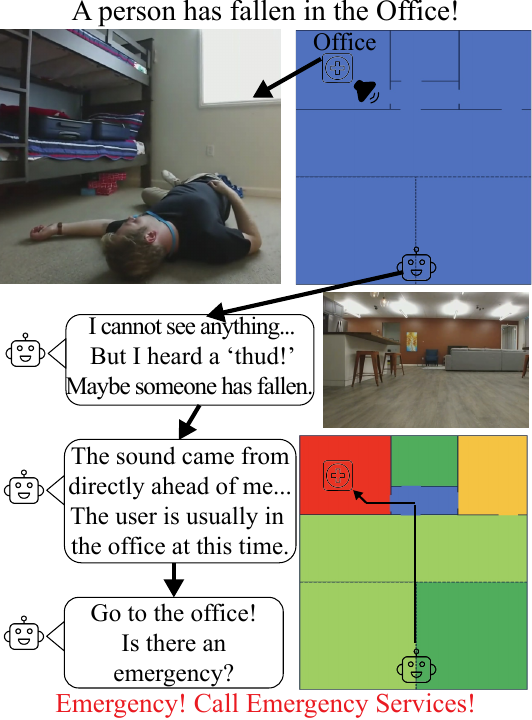}
     \vspace{-0.2cm}
     \caption{An overview of our method. Our agent hears a "thud" and determines that a fall may have occurred. It then leverages the probabilistic edges of our probabilistic dynamic scene graph (P-DSG), representing a heatmap of agent activity, and audio direction to update our P-DSG and produce a hypothesized source location. In this example, the P-DSG shows a high probability of the user, and thus any fall, being in the office (the red box). The agent checks the office and detects a fallen person. It calls emergency services.
     }
     \label{fig:coverimg}
    \vspace{-0.7cm}
 \end{figure}

\IEEEPARstart{I}{magine} you have an elderly parent at home and they take a hard fall. This situation is very serious and potentially deadly, with a quick emergency response essential to a positive outcome. In the United States in 2022 accidental home deaths exceeded 128,000, over three times the death rate of motor vehicles \cite{mackUnintentionalInjuries2013}. 34\% of these deaths are due to falls directly with another 48\% potentially leading to falls (poisoning or choking) \cite{DeathsHomeIntroduction}. Additionally, almost 7 million people in the United States were treated in emergency rooms for fall or trip related injuries in 2020 alone. Fires represent the next largest cause of death and damages, accounting for 10\% of accidental home deaths and over \$10 Billion in damages each year \cite{FireLossUnited2023}.

Now imagine a home robot that can monitor the home and respond to these types of emergencies quickly. Their ability to contact emergency services in a timely fashion would make a serious difference in outcome \cite{adeyemiEmergencyMedicalService2023}.

As a home robot cannot be present in multiple rooms at once, audio is a natural cue to detect possible emergencies in the home. The audio signal can be perceived the instant something occurs, enabling a speedy response. It also contains information about what may be occurring, like a `thud' noise for a hard fall, or an alarm for a fire.

The AudioGoal \cite{soundspaces} task introduced Audio-Visual (AV) navigation with a goal of finding the source of a periodic (continuous) sound using egocentric images and captured binaural audio. This is similar to the ObjectGoal task but with audio instead of a target object. Subsequent works added shorter duration sounds, and semantic grounding of sounds \cite{chen_semantic_2021, gan_finding_2022}.
For emergencies in a home environment, these tasks are inadequate. 
Additionally, no known tasks allow prior exploration of the scene, which a home robot will have.

\textbf{Main contributions.} We introduce the HomeEmergency dataset to benchmark the abilities of embodied AI agents to identify emergencies in the scene. Figure \ref{fig:coverimg} showcases our task. When initialized, an agent is spawned into the simulator and has the opportunity to explore the environment as a home robot would in the real world. At runtime, the robot is again placed into the simulator at a random point, but this time an audio signal is played in the environment representing a real emergency (a fallen person) or a false emergency (a fallen box or suitcase). The agent must efficiently find the source of the audio and determine if an emergency is occurring.

The HomeEmergency task is challenging due to its aperiodic sounds, multiple rooms, and partial observability of the environment. Additionally, agents must not just \textit{find} the source of the audio, but effectively \textit{determine whether an emergency has or has not occurred} when presented with a true emergency or false emergency.  

We propose a modular method for this task consisting of perception, mapping, fusion, and emergency identification modules. These modules use the audio propagated through the scene to determine that an emergency may be occurring, before using the direction the audio came from and a probabilistic dynamic scene graph generated through previous exploration of the scene to determine the likeliest location of the audio source. The agent navigates to the audio source and uses the emergency identification module to determine if an emergency is occurring. 

Our main contributions are as follows:
\begin{enumerate}
    \item The HomeEmergency dataset, aimed at enabling researchers to create embodied agents capable of responding to emergency situations in the home. 
    We provide code for users to generate data samples.
    \item A modular method for the HomeEmergencySim task, consisting of four components: perception, mapping, fusion, and emergency identification.
    \item A novel scene graph structure which extends dynamic scene graphs to mobile objects or agents by using probabilistic edges, which represent the likelihood of the agent being in a given location.
    \item A novel inference algorithm that uses Bayesian inference to update the probabilistic dynamic scene graph with new information and hypothesize the location of an agent for quicker response to emergencies.
    \item We conduct a sim-to-real transfer of our method and demonstrate its performance.
\end{enumerate}



%% file: Sections/2-Related.tex
\section{Related Work}
\subsection{Audio-Visual Navigation}
Audio-Visual (AV) navigation was introduced in AudioGoal \cite{soundspaces} where the agent has to navigate to the source of a periodic sound using egocentric images and captured binaural audio. Initial work focused on navigating to the sound source by inferring the source of the sound and moving towards it \cite{gan2020look, deitke_retrospectives_2022}. Chen et al. \cite{chen_learning_2020} builds upon these works by using reinforcement learning, setting waypoints, and mapping the audio to the scene. 
Subsequent research has brought AudioGoal closer to the real world. Chen et al. (SemanticAudioGoal) \cite{chen_semantic_2021} limits sound length and localizes sounds to an object that makes semantic sense. Gan et al. \cite{gan_finding_2022} (FFO) has the agent find fallen objects with realistic collision sounds. Additional research has extended these tasks to multiple goals \cite{kondoh_multi-goal_2023} and to real world agents \cite{chen_sim2real_2024}.

Our task expands upon these works by 
utilizing a series of real-world sounds, propagated as they would in the real-world, of the same length as in the real-world, and coming from a sounding object that makes semantic sense. Our task includes periodic and aperiodic sounds, as well as many emitted by humans, which to our understanding has not been in any AV navigation task. 

\subsection{Scene Graphs}
Scene graphs are a common method of representing a scene in computer graphics and 3D modeling where, generally, nodes of the graph are objects and edges are relationships. 
Creating a scene graph from images is a popular problem in the computer vision community \cite{yang2018graph, xu2017scene}. Many robotics simulation platforms are built on top of a scene graph including both Habitat \cite{savvaHabitatPlatformEmbodied2019} and VirtualHome \cite{puig2018virtualhome}.

Dynamic Scene Graphs (DSG) build off of this work and allow for objects or agents to move. \cite{rosinol3DDynamicScene2020, rosinolKimeraSLAMSpatial2021} first proposed DSGs alongside methods to create the graph and track agents to update the graph. Subsequent work by Gorlo et al. \cite{gorloLongTermHumanTrajectory2024} utilized DSGs for predicting future human trajectories. Different from these works, which assume a fully-observable environment for updating the graph, we propose a DSG with informed probabilistic edges, which we call the probabilistic dynamic scene graph (P-DSG) that can connect one agent to many potential locations in the scene.

\subsection{Language-Based Embodied AI}
Leveraging language to inform a robotic agent is a well studied task in the literature. Formative work included using generalized grounding graphs \cite{tellex2} for robot manipulation \cite{ggs}, performing language-guided navigation \cite{safenav, robotrust}, autonomous driving \cite{langdynamic}, and drone-control \cite{dronelang1}.
Tellex et al. \cite{tellex1} recently presented a survey on using language for robots.

Recent work tackling this problem by Thomason et al. \cite{jesse1, cvdn} and Gao et al has explored the use of human-robot dialogue to gather relevant information for completing tasks. DialFRED \cite{dialfred} explores utilizing human-robot dialogue to ascertain relevant information for completing tasks. 
Dorbala et al. \cite{dorbala} uses LLMs for object goal navigation. 
Mullen et al. \cite{mullenAnomaly} expands upon these works by using natural language derived from a scene graph to detect many dangers in the environment. We build upon these works by using language to both inform our P-DSG scene representation with object characteristics (e.g. likelihood of causing a fire) and visually verify if an emergency is occurring.

%% file: Sections/3-Dataset.tex
\section{The HomeEmergency Dataset}
\subsection{Creating the HomeEmergency Dataset}

 \begin{figure}[t]
     \centering
     \vspace{0.2cm}
     \includegraphics[width= \linewidth]{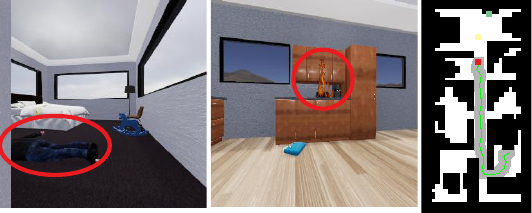}
     \vspace{-0.6cm}
     \caption{A sample of images from the simulator showing simulated fall emergencies (left) and fire emergencies (middle). A 2D occupancy map showcasing the complexity of the overall environment is to the right. It also shows our method proceeding towards a fire, faded green dot, along a very efficient path, bright green.
     }
     \label{fig:sim_ims}
     \vspace{-0.7cm}
 \end{figure}
 
 \begin{figure*}[t]
     \centering
     \includegraphics[width= 0.8\linewidth]{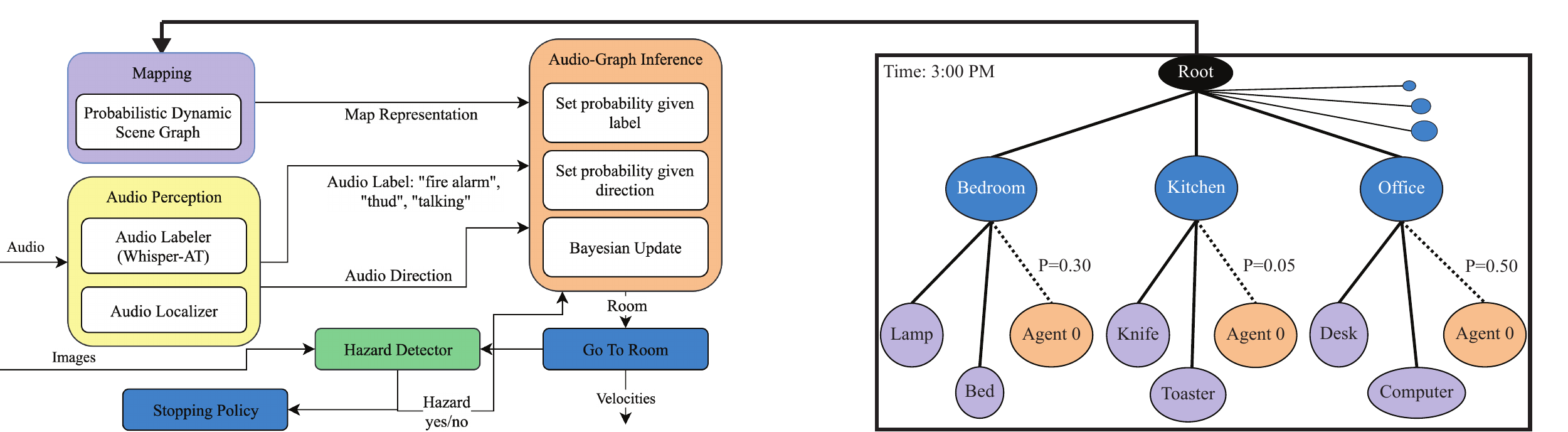}
     \caption{An overview of our modular method and our mapping representation, the P-DSG. Our method takes inputs of audio and images. The method begins with the \textit{Mapping} module which creates a probabilistic dynamic scene graph (P-DSG) (right).
     The audio is run through our \textit{Audio Perception} module which outputs a label of the audio and an estimated direction it comes from. In our \textit{Inference} module, we use this perception information and information from our P-DSG to determine the most likely room to be the source of the audio. We then go to this room, checking for emergencies, and updating our probabilities accordingly. The method continues until an emergency is found, the house is cleared, or in simulation, we run out of steps. To the right is our P-DSG where objects and agents are connected to their parent place and room with edges (places left out for simplicity). Note the probabilistic edges for the dynamic agents.
     }
     \label{fig:method}
     \vspace{-0.2cm}
 \end{figure*}

\textbf{Task Definition.} In HomeEmergency, an embodied agent is tasked with using audio to detect a potential emergency situation in the home. Upon hearing audio which is likely connected to an emergency, the agent must navigate to the source of the audio and determine if there is an emergency. Before running a scenario, the agent is provided the opportunity to explore the scene as a real home robot would; a key difference between our dataset and previous work which assumes a zero-shot environment \cite{soundspaces, chen_semantic_2021, gan_finding_2022}. A 2-D occupancy map of free space and occupied space, a 2-D map of room definitions, and a room-based heatmap of human activity with added Gaussian noise is provided to the agent. In the real world, it is assumed that the agent can develop this model of human activity over time.

\textbf{Simulator and Scene.} We use ThreeDWorld (TDW)\cite{ganThreeDWorldPlatformInteractive2021} as the simulator underpinning the HomeEmergency dataset. TDW was chosen for its inclusion of audio propagation, extensive documentation, humanoid avatars, and native robotic agent platform. TDW also provides 36 unique multi-room home scenes which we utilize for this version of our dataset. These scenes are photorealistic and include on average 7 distinct rooms of common room types like kitchen, bedroom, living rooms, and offices. The area of these scenes is generally between 1000 and 1500 square feet. Note that TDW allows for the creation of new unique floorplans for future iterations of the HomeEmergency dataset.

\textbf{Adding Emergencies.}
In HomeEmergency we choose to represent falls and fires. Falls directly represent 34\% of home deaths and potentially occur in another 48\% of deaths (poisoning and choking). Fires are the next most common source of home deaths and property losses with actionable audio. 
As not every fire alarm noise will be connected with a fire and not every fall noise will be connected with a human falling, we include samples affiliated with each class where there is not an emergency. For example, a fall noise which prompts the agent to find a potential fallen person may have come from a box falling off of a shelf. False positives can inundate emergency services and make the robot less useful, therefore the agent must not report these instances.

For falls, we determine where to place the human in the scene by using HOMER \cite{patelProactiveRobotAssistance2022}. HOMER generates a schedule of a sample human agent's activities throughout a day. We select a random time and run HOMER until we get the human activity for this time, for example 18:00 and `cooking dinner.' We then map this activity to which room the activity is likely to occur in (the kitchen) then place the human in this room. Every fall episode uses aperiodic audio, a fall sound followed by no additional sounds. 

For fires, we affiliate the fire with a source which makes semantic sense, like a skillet or a toaster. We randomly choose between all potential sources available in the scene. Different from falls, we allow fires to produce periodic sound - continuous audio like that from a fire alarm, aperiodic noise - as if a fire alarm starts and then stops quickly, or semiperiodic noise - as if the alarm starts, followed partway through the episode by either stopping or additional fire alarms rendering new audio direction information worthless.

For all `negative' data samples, those with a fall or fire related noise but do not present an emergency, the source of the audio is placed in a random location in the scene. 


\subsection{Using the HomeEmergency Dataset}\label{procedures}

\textbf{Episodes.} Each episode in the HomeEmergency dataset begins by instantiating one of the 36 scenes in the ThreeDWorld simulator. The agent is then spawned into the scene at a random location before audio is added into the scene at the source and propagated through the scene using the simulator's physics engine. The agent is provided with a 2D occupancy map, a 2D room map, and a noise-added human activity heatmap. The heatmap is created by using the HOMER-created activity-based heatmap and mapping it to rooms. To model likely imperfections in any estimation of human activity, we add Gaussian noise with $\sigma = 0.05$ to the heatmap before providing it to the agent. 

\textbf{Agent.} ThreeDWorld contains a robotic agent called the Magnebot, a large mobile robot with an attached RGBD camera and microphone for egocentric perception. The Magnebot is moved through the scene in a discrete manner with commands for turning 15 degrees right or left, 0.15 meters forward, and collecting new audio. 

\textbf{Evaluations.} We evaluate agents for both effectiveness and efficiency. 
The key metrics we utilize are:
\begin{enumerate}
    \item \textbf{AudioGoal Success Rate (AG SR)}: The fraction of episodes where the agent navigates to and visualizes the audio source. We use oracle stopping for all methods with this metric.
    \item \textbf{AudioGoal Success Rate weighted by Inverse Path Length (AG SPL)}: $AG SPL = AG SR\*(PL/oPL)$ where $PL$ is path length and $oPL$ is optimal path length. Inefficient, paths are penalized in this metric.
    \item \textbf{Emergency Detection False Negative Rate (EDFNR)}: The fraction of episodes that do not report an emergency when there is one.
    \item \textbf{Emergency Detection False Positive Rate (EDFPR)}: The fraction of episodes that report an emergency when there is not one. 
\end{enumerate}
As existing AG methods do not provide a mechanism for detecting emergencies, we utilize the AudioGoal SR and SPL metrics to characterize each methods performance in finding the source of the audio. The EDFNR and EDFPR are more illustrative metrics for our dataset, characterizing failures to properly report or not report an emergency.

%% file: Sections/4-Method.tex
\section{Detecting Emergencies in the Home}
\begin{table}[t] 
    \centering
        \caption{List of common symbols used and their definitions.}
    \label{tab:symbols}
    \begin{tabular}{| p{0.14\columnwidth} | p{0.76\columnwidth} | } 
        \hline
        \textbf{Symbols} & \textbf{Definitions}\\  
        \hline
        $\mathcal{P}(E)$ & The probability of an emergency in the scene.\\
        \hline
        $a$ & The audio observation in a given episode. $a$ has two components: $a_l$, audio class, and $a_{dir}$, audio direction. \\
        \hline
        $G$ & The scene representation prior, for our method the P-DSG. \\ 
        \hline
        $o_o, o_{ag}$ & Nodes on the object layer of our P-DSG. They represent static objects and dynamic objects/agents respectively.\\
        \hline
        $S, S^*$ & The current scene and the partially-observable scene which can be navigated to ascertain new information. \\
        \hline
        $\mathcal{P}(Ag|r)$ & The probability of an agent, given a room in the home. \\
        \hline
    \end{tabular}

     \vspace{-0.6cm}
\end{table}
A list of symbols frequently used in this work is shown in Table \ref{tab:symbols}. Rarely used symbols are defined where they are used. We address the task of using audio, $a$, to detect an emergency situation in the home, before navigating to the audio source and determining if the source is an emergency. 
The goal of our task is to evaluate the following equation:
\begin{equation} \label{eq:main}
    \mathcal{P}(E|a, G, S)
\end{equation}
and report the emergency if there is one. Agents must navigate the partially-observable scene, $S^*$,  and infer relevant information from it.

We present an overview of our modular method for this task in Figure \ref{fig:method}. 

\subsection{Mapping Representation: P-DSG}
Before an episode we explore the scene and create a prior to inform our subsequent response to potential emergencies. To encapsulate our prior we propose a Probabilistic 3D Dynamic Scene Graph (P-DSG). The base dynamic scene graph (DSG) representation provides a hierarchical symbolic extraction of the environment formally consisting of nodes and edges, $G = \{V,E\}$. Following existing literature we structure the DSG into three layers, rooms, $r$, places, $p$, and objects, $o$. In this structure, rooms are areas like a kitchen or bedroom, places represent the free space in the environment, and objects denote a semantic object. 

We extend this structure for our method with our `objects' layer containing two types of nodes, $o_o$ and $o_{ag}$. In our implementation, $o_o$ nodes consist of semantic labels as in many existing scene graph methods. We believe our P-DSG should work seamlessly with more flexible implementations with open-vocabularies like VLPG \cite{arul2024VLPG} or CLIO \cite{Maggio2024Clio}
which leverage CLIP \cite{CLIP} embeddings 
to build the graph
Traits, like an objects likelihood of causing a fire, causing a trip, or danger to children, can be added onto the $o_o$ nodes using LLMs for later querying. $o_o$ nodes allow us to locate static objects in the scene, for example if we hear a fire alarm, we can find fire alarms or objects likely to cause fires, like a stove. $o_{ag}$ nodes are affiliated with dynamic objects or agents like humans or pets. As agents can move, these nodes are probabilistic, containing an estimated probability of the agent being in a given place or room. In practice, these probabilities are set by learning the agent’s routine, and should be adjusted in real time based off of the current visual observations of the scene, $S^*$. For example, if a user typically eats dinner at 6:00, the base probability of them being in the dining room at 6:00 would be high. However, if we navigate through the kitchen at 6:00 and they are not there, we lower the probability accordingly. 
Note that our environment is partially observable and we cannot know where the user is at all times. 


\subsection{Audio Perception}
At task run time, the agent uses audio to detect that an emergency situation may be occurring. Specifically, we employ Whisper-AT to label all incoming audio captured from a simulated microphone on the agent in the scene, at overlapping 4 second intervals. The labels provided from Whisper-AT are the top similarity from a set of 527 possible labels. If the label for an audio clip, $a_l$, is from a predetermined list deemed to be affiliated with a potential emergency situation, like ‘thud’ or `thunk' for falls, it will trigger the agent to localize the source of the audio. We also use the direction of the audio source, $a_{dir}$. If the audio is periodic, audio direction and labels get updated accordingly.


\subsection{Inference}
To solve Equation (\ref{eq:main}), we must use $S^*$ to more fully understand the current state of the scene. We assert that by finding the source of the audio, we can use vision (and our emergency identification module) to optimally update $\mathcal{P}(E)$. As such, finding the probability that each room is the source of the audio, $\mathcal{P}(r | a)$, becomes the intermediate task. We apply Bayesian inference to evaluate this equation with respect to the perceived audio:

\begin{equation}\label{eq:bayes}
    \mathcal{P}(r_i|a) \propto \mathcal{P}(r_i)\mathcal{P}(a|r_i).
\end{equation}

Initially, we set the prior, $\mathcal{P}(r_i)$ to be equal for all rooms in the scene. We evaluate $\mathcal{P}(a|r_i)$ by breaking $a$ into its component parts, $a_{dir}$, and $a_l$. This gives us

\begin{equation}
    \mathcal{P}(a|r_i) = \mathcal{P}(a_l|r_i)\mathcal{P}(a_{dir}|r_i).
\end{equation}

Our P-DSG, $G$, enables us to evaluate $\mathcal{P}(a_l|r_i)$. For the HomeEmergencySim task, we map audio labels we ground in falls to leverage $o_{ag}$ nodes and those we ground in fires to leverage $o_{o}$ nodes. 
For instances of $a_l$ which we ground in agents, we assert $P(a_l|r_i)\approx P(Ag|r_i)$ as any emergency is coming from an agent.
As such, we estimate $P(Ag|r)$ directly from the probabilistic edges of the $o_{ag}$ nodes.

For instances of $a_l$ which we ground in objects, we evaluate $\mathcal{P}(a_l|r_i)$ by determining which objects and subsequent rooms are likely to cause $a_l$. For example, in the case of an $a_l$ of `fire alarm,' we can query our $o_o$ nodes for objects likely to cause fires. Specifically in our method, we evaluate the probability of each $o_o$ node to cause a fire using a prompt to an LLM 
and establish this as a trait of the object. After querying each $o_o$ node for this trait we sum the probabilities for each room before normalizing to give us $\mathcal{P}(a_l|r_i)$.

To evaluate $\mathcal{P}(a_{dir}|r_i)$, we take the path from the robot to the room, $r_i$ in the scene and determine if the perceived audio direction follows this path initially. We evaluate this as a step function with rooms whose path intersects with $a_{dir}$ (within a threshold) given $\mathcal{P} = 0.99$ and those who do not given $\mathcal{P} = 0.01$. Intuitively, if the agent is in a room with a door in front of and behind it, and it hears a sound in front of it, all rooms accessible from the forward direction would be a possible source of the sound but all the others would be essentially eliminated. In practice, the threshold and step probabilities can be adjusted to account for errors or noise in the sound direction measurements.

After fully evaluating equation \ref{eq:bayes}, we update the P-DSG probabilities as necessary, before taking the room with the highest probability and begin navigation towards it. If new audio is captured, we recalculate $\mathcal{P}(a|r_i)$. 

As evaluating $\mathcal{P}(a|r_i)$ is a fiat for evaluating our ultimate goal, $\mathcal{P}(E)$, we continuously run our emergency identification module and update the prior $\mathcal{P}(r_i)$.
Intuitively, if we do not spot the emergency in $r_i$, we adjust $\mathcal{P}(r_i)$ such that we continue searching other rooms. In negative episodes, where the audio was not created by a emergency, like a fallen box, this will cause our agent to continue to navigate until the entire scene is cleared of potential emergencies.
We believe this is preferable behavior to prevent false negatives and a delayed emergency response.

\subsection{Emergency Identification}
Throughout each episode, we collect egocentric RGB images and attempt to detect a possible emergency. To identify potential emergencies from the agent's POV, we use the multimodal LLM LLaVA 1.6 \cite{liu2024llavanext} and feed in the egocentric image from the agent as it navigates the environment, as well as a 360 degree image when entering a new room, akin to a human quickly glancing left and right. We prompt LLaVA to identify a potential emergency with the following prompt:

\begin{quote}\textit{
    You are a home robot whose job is to make sure your human users are safe in their home, preventing or responding to common sources of deaths in the home. In an image, you must check for anything that may constitute an emergency situation for the user. For example, if the user is on the floor, they may have taken a fall. This constitutes a potential emergency situation and we must check on the user. Additionally, seeing a fire is a clear danger and emergency services should be contacted. Are there any potential fallen users, dangers, or active emergency situations in the provided image? Please state yes or no. You then HAVE TO explain your reasoning.}
\end{quote}

%% file: Sections/5-Experiments.tex
\section{Experiments and Results}
\subsection{Implementation Details}
Experiments are run following the described setup from \hyperref[procedures]{Section III.B }. We specifically run 1152 episodes split evenly between fall and fire emergencies, and between positive and negative samples. Note that essentially infinite samples could be generated but we chose this amount to match the scale of similar prior art. In each episode, the agent is spawned into the scene at a location and orientation defined by the episode, with audio then propagated through the scene to the agent using the simulator's physics engine. The simulator also provides a 2D occupancy map, 2D room map, and noise-added human heatmap to the agent at the beginning of the episode. The agent then has a limit of 500 discrete steps in the environment to find the potential emergency. This both follows exiting literature and allows us to distinguish performance between methods as otherwise they would continue exploring the entire scene until they encounter the hazard, resulting in 100\% AG SR for all methods.

For getting the audio direction in the simulator, due to limitations in audio perception with multiple microphones, we utilize a pseudo-truth audio direction by providing the direction of the nearest room opening boundary along the shortest path from the agent to the audio source. This represents that in our multi-room scenario, if the agent is not in the room with the audio source, the audio will propagate through the scene to the agent through the doorway. In the real world, we get audio direction as the orientation of the microphone with the highest amplitude on an array of mics. While this works well for our use cases, more sophisticated methods of determining audio direction exist and can be utilized with our method. An audio recording captured from the Agent's POV is used for determining the audio label.

We use a self-stopping policy for our task Emergency Detection metrics where the agent only stops if it detects an emergency.
We report the performance for falls and fires separately for easier evaluation. This is especially valuable as falls represent largely aperiodic sounds and is an emergency \textit{directly} connected to a human. Conversely fire sounds can represent themselves as aperiodic or periodic, and are not always caused by humans, with many fires starting due to faults in appliances or electrical equipment.

\vspace{-0.5cm}
\subsection{Baselines and Ablations}
We consider 3 baseline methods for evaluations:
\begin{itemize}
    \item \textbf{Direction Following (DF) \cite{gan2020look}}: This is a privileged version of the method from \cite{gan2020look} with pseudo-ground truth audio direction information. The method uses the geography of the local environment and audio direction to navigate towards the audio source.
    \item \textbf{Finding Fallen Objects (FFO) \cite{gan_finding_2022}}: This is a direct application of the FFO method from \cite{gan_finding_2022} as trained in the TDW simulator. It uses a neural network to determine the hypothesized source of the audio event. 
    \item \textbf{PPO \cite{schulmanProximalPolicyOptimization2017}}: Similar to previous work, we train an end-to-end RL policy using Proximal Policy Optimization (PPO) and maximize the reward of finding the audio.
    This model takes the input history of RGBD images as well as audio observations, the 2D occupancy map, 2D room map, noise-added human heatmap, and outputs actions that the agent executes in the environment.
\end{itemize}
We selected these three baselines as they are both commonly used baselines in AudioGoal literature, and the most adaptable to the unique constraints of our task. We believe they represent the \textit{existing} state-of-the-art on our task. All methods utilize the 2D occupancy map provided from the episode. As FFO does not use audio direction, instead relying on a trained network to provide an estimated audio location, it does not utilize our pseudo-truth audio direction. PPO is the only baseline that utilizes the noise-added human heatmap and 2D room map as the others do not naturally utilize this information and extending them to rely on it was not feasible.
All baselines are run with our \textit{Emergency Identification} module for collecting emergency detection metric results. They do not have this ability natively.  

To train the PPO method, following prior literature \cite{gan_finding_2022}, we set up the reward such that at each step the agent receives a reward of +1 if it is closer to the target location and -1 if it is further from it. The agent receives a reward of +10 if it finds the audio source. There is also a -0.01 penalty for each time step. We use CNNs to encode the RGBD images, the 2D occupancy map, and the audio spectrogram at each time step as a feature vector and concatenate them with the room and heatmap information as an input for the Gated Recurrent Unit (GRU) model. We train the model until convergence.

We also ablate different components of our method to test how they effect performance individually.
\begin{itemize}
    \item \textbf{Ours w/o $\mathcal{P}(a_{dir}|r_i)$}: This completes our \textit{inference} module without updates related to audio direction, instead relying solely on the audio label to inform our hypothesized audio source location.
    \item \textbf{Ours w/o $\mathcal{P}(a_{l}|r_i)$}: This completes our \textit{inference} module without updates related to audio class. This means that the P-DSG, including the agent heatmap and object semantics, is not used to inform our hypothesized audio source location.
\end{itemize}

\subsection{Comparisons with Baselines}

\begin{threeparttable}[t]
        \caption{Results for the `Falls' class of the HomeEmergency task.}
    \centering
    \begin{tabular}{l|c c c c} 
        Method & AG SR $\uparrow$ & AG SPL $\uparrow$ & EDFNR $\downarrow$\\ [0.5ex] 
        \hline\hline
        DF \cite{gan2020look} & 0.57 & 0.52 & -/0.53 \\
        \hline
        FFO \cite{gan_finding_2022} & 0.17 & 0.15 & -/0.80\\
        \hline
        PPO \cite{schulmanProximalPolicyOptimization2017} & 0.11 & 0.09 & -/0.91\\
        \hline\hline
        \textbf{Ours} & \textbf{0.75} & \textbf{0.63} & \textbf{0.20}\\
        \hline\hline
        Ours w/o $\mathcal{P}(a_{dir}|r_i)$ & 0.58 & 0.47 & 0.47 \\
        \hline
        Ours w/o $\mathcal{P}(a_l|r_i)$ & 0.58 & 0.51 & 0.45\\
      \hline\hline
        Ours w/ periodic  & 0.90 & 0.72 & 0.07 \\
        \hline
    \end{tabular}
    \begin{tablenotes}
        \small {
        \item {Note: We significantly outperform all baselines (top section). We also show the importance of our $\mathcal{P}(a_{dir}|r_i)$ and $\mathcal{P}(a_l|r_i)$ inferences. Notice the bottom line, showing performance with periodic fall audio, showing both the difficulty of our task and the performance of our method when provided with continued information.}
        }
    \end{tablenotes}
    \label{tab:fall-results}
    \vspace{0.2cm}
\end{threeparttable}

\begin{threeparttable}[t] 
    \caption{Results for the `Fires' class of the HomeEmergency task.}
    \centering
    \begin{tabular}{l|c c c c} 
        Method & AG SR $\uparrow$ & AG SPL $\uparrow$ & EDFNR $\downarrow$\\ [0.5ex] 
        \hline\hline
        DF \cite{gan2020look} & 0.72 & 0.67 & -/0.47 \\
        \hline
        FFO \cite{gan_finding_2022} & 0.11 & 0.11 & -/0.89\\
        \hline
        PPO \cite{schulmanProximalPolicyOptimization2017} & 0.14 & 0.13 & -/0.92\\
        \hline\hline
        \textbf{Ours} & \textbf{0.86} & \textbf{0.73} & \textbf{0.31}\\
        \hline\hline
        Ours w/o $\mathcal{P}(a_{dir}|r_i)$ & 0.68 & 0.61 & 0.47 \\
        \hline
        Ours w/o $\mathcal{P}(a_l|r_i)$ & 0.80 & 0.71 & 0.39\\
        \hline
    \end{tabular}
    \begin{tablenotes}
    \small {
    \item {Note: We significantly outperform all baselines (top section). We also show the importance of our $\mathcal{P}(a_{dir}|r_i)$ and $\mathcal{P}(a_l|r_i)$ inferences (bottom section).}
    }
    \end{tablenotes}
    \label{tab:fire-results}
    \vspace{0.2cm}
\end{threeparttable}

In table \ref{tab:fall-results} we report the overall performance of our agent, baselines, and ablations on the `falls' class of our dataset. Our full method performs the best on this task with a 32\% improvement over the best baseline, \textbf{DF} \cite{gan2020look}, in AG SR and 62\% improvement in emergency detection false negative rate (EDFNR). The \textbf{FFO} and \textbf{PPO} trained methods both exhibit serious failures. We observe that \textbf{FFO} fails in the multi-room, aperiodic, setting, likely as training the model to predict the audio source location through multiple rooms is intractable. We also observe that the RL model (\textbf{PPO}) performs poorly on our complex task. Notable is that our method performs even stronger on episodes with a \textit{true positive} fall with an AG SR of 0.80. This is likely due to our method prioritizing finding a potential emergency over finding the source of the sound, proceeding towards rooms where encountering a human is more likely and potentially skipping over low probability rooms where the distractor (e.g. suitcase falling) sound originated from. 
Our \textit{emergency identification} module produces very few false positives, with an EDFPR of 0.03.

We also present our results on `fires' in Table \ref{tab:fire-results}. Audio direction is even more important for this class as the audio is frequently periodic, providing more information than the instantaneous sounds from the `falls' class. We continue to exhibit a 19\% improvement over \textbf{DF} despite its own stronger performance.
\textbf{FFO} and \textbf{PPO} continue to perform poorly.

Task failures which contribute to the EDFNR for our method fall into three main categories, 1) suboptimal navigation/running out of navigation steps (74\% of failures), 2) colliding with a wall or object in the scene causing a simulation failure (21\% of failures), and 3) the emergency identification module VLM producing a false negative result directly ($<$5\% of failures). Suboptimal navigation is largely characterized by AG SR, and is a direct result of not going to the correct location quickly enough. Failure mode 3 is caused almost exclusively by a poor view of the target.

\subsection{Comparisons with Ablations}

In Tables \ref{tab:fall-results} and \ref{tab:fire-results} we show our results against our ablations. Removing the $\mathcal{P}(a_{dir}|r_i)$ and $\mathcal{P}(a_l|r_i)$ updates result in similar decreases in performance, with $a_{dir}$ resulting in a stronger decrease in SPL performance. This makes sense as the lack of direction would be likely to cause the method to pursue exploration of rooms in the wrong direction, increasing path length. Both methods result in a 29\% decrease in AG SR and a 2.3x increase in the EDFNR for falls. Notably, the ablation without $\mathcal{P}(a_l|r_i)$ is similar to the \textbf{DF} baseline and exhibits similar AG SR performance. However, the EDFNR is still lower than \textbf{DF}, which we can attribute in part to our P-DSG allowing us to clear rooms of potential emergencies more quickly. For fires, removing the $\mathcal{P}(a_l|r_i)$ inference update still decreases performance, but not as significantly as for `falls,' largely due to the periodic audio providing more information to the $a_{dir}$ updates.

\subsection{Evaluations of HomeEmergency}
Table \ref{tab:fall-results}, we also show our results should falls result in periodic sounds, like a user potentially calling for help after they've fallen down. This illustrates some of the difficulty of estimating the location of a fallen person with only instantaneous sound, as well as the current performance ceiling for our method, with the agent achieving an AG SR of 0.9 and EDFNR of only 0.07.

\begin{threeparttable}[t]
    \centering
    \caption{Ablation results for amount of noise in the provided human heatmap for the `Falls' class of the HomeEmergency task. }
    \begin{tabular}{l|c c c c} 
        Method & AG SR $\uparrow$ & AG SPL $\uparrow$ & EDFNR $\downarrow$\\ [0.5ex] 
        \hline\hline
        \textbf{Ours} & \textbf{0.75} & \textbf{0.63} & \textbf{0.20} \\
        \hline\hline
        Ours w/ $\sigma = 0.05$  & 0.73 & 0.60 & 0.23 \\
        \hline\hline
        Ours w/ $\sigma = 0.15$ & 0.67 & 0.55 & 0.37 \\
        \hline
        Ours w/ $\sigma = 1.0$ & 0.62 & 0.52 & 0.43 \\
        \hline
    \end{tabular}
    
    \label{tab:fall-noise-abl}
    \begin{tablenotes}
    \small {
    \item {Note: A small amount of added noise does not severely disadvantage our method, but significant added noise brings performance close to that without the use of the heatmap.
    }}
    \end{tablenotes}
    \vspace{0.2cm}
\end{threeparttable}

With the human heatmap providing significant information to our P-DSG, and in turn our \textit{inference} module, we ablated against the amount of noise in the map. For this, we took the episode provided heatmap, which already added noise with $\sigma = 0.05$, and add additional noise with $\sigma = 0.05$, $0.15$, and $1.0$. We show that our method is relatively robust to additional added noise with $\sigma = 0.05$. Additional noise from there continues to degrade performance.

\subsection{Real World Demonstration}
To demonstrate the ability of our method to transfer to the real world, and the applicability of our HomeEmergency Dataset, we conduct a series of 30 real world experiments replicating the falls class of the HomeEmegency Dataset. These experiments are split evenly between positive samples (fallen human) and negative samples (fallen other object). To closely match the HomeEmergency Dataset we use HOMER for positive samples to assign the human to a room and to gather a noise-added human heatmap for the Agent. We utilize 4 single-floor apartment environments (between 900 and 1400 square feet) in our experiments. We acknowledge that this scenario is a simplified version of the real world which may include multiple floors, additional rooms, novel objects, and significant background noise. These experiments are intended as a proof of concept that the sim2real transfer of our method can work in these simplified scenarios.

We put our method on a small mobile robot with an egocentric camera and an array of microphones. Our method is used as is, with the robot first creating a P-DSG graph of the scene. For each scenario, the robot must use its own sensors to collect audio information and validate a potential emergency. We provide a human activity heatmap. 

{\centering    
\begin{threeparttable}[t] 
\centering
\caption{Real World Testing Results.}
\begin{tabular}{l|c c} 
    Method & AG SR $\uparrow$ & EDFNR $\downarrow$\\ [0.5ex] 
    \hline\hline
    DF \cite{gan2020look} & 0.60 & -/0.47 \\
    \hline
    \textbf{Ours} & \textbf{0.83} & \textbf{0.10}\\
    \hline
\end{tabular}
\label{tab:rw-results}
\vspace{0.2cm}
\end{threeparttable}}

We find that performance is similar between the simulator and the real world. We notice an increase in performance related to navigation, with fewer navigation failures and simpler layouts than in the simulator, alongside a drop in performance which we attribute to additional noise in the audio direction. Our method mitigates this in part by navigating based on rooms, while DF is impacted more severely. Similar to in simulation, running out of steps caused the most failures in our method. This occurred more in negative samples, where the agent explores high probability rooms and neglects the source room. There were no false positives or negatives connected to our VLM based emergency detection module.

We want to highlight three common cases from our testing:
\begin{enumerate}
    \item \textbf{A human falls close by.} The close room may be a low probability room for the user to be in, but is in a unique direction.
    \item \textbf{A human falls far away.} The far room is a high probability room for the user to be in, but audio direction leads to many potential rooms.
    \item \textbf{An object falls far away.} The same situation as the previous, but no human has fallen. The agent must clear the home of potential emergencies.
\end{enumerate}

  \begin{figure}[t]
     \centering
     \vspace{0.2cm}
     \includegraphics[width= 0.6\linewidth]{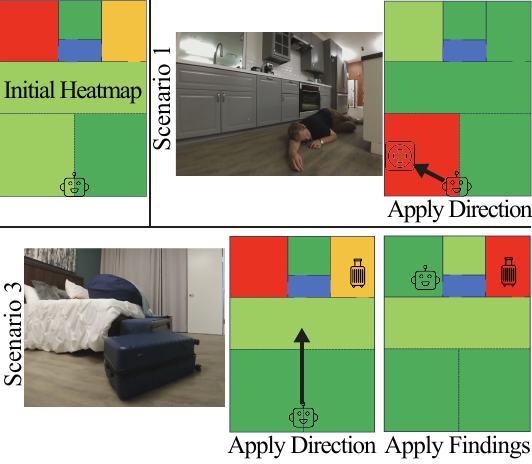}
     \vspace{-0.2cm}
     \caption{We show how out method responds to the scenarios in our real world testing. The initial heatmap used for all scenarios is in the top left. 
     }
     \label{fig:real_exp}
     \vspace{-0.6cm}
 \end{figure}

In all three cases our method exhibits the intended behavior. Illustrated in Figure \ref{fig:real_exp}, for scenario 1, our \textit{inference} module selects the kitchen as the most likely source of a potential emergency. The robot navigates there and our \textit{emergency identification} module properly identifies the fallen human. For scenario 2, Illustrated in Figure \ref{fig:coverimg}, utilizing the human heatmap, the robot navigates to the office first, finding the fallen human quickly. In scenario 3, the robot again navigates to the office, thinking there may be a fallen human there, before continuing to the bedroom, where it identifies the suitcase on the floor. The remainder of the home is still checked for falls where none are encountered. 

%% file: Sections/6-Conclusion.tex
\section{Conclusions, Limitations, and Future Work}

In this work we share a new task, HomeEmergency, for enabling robotic agents to respond to emergency situations in the home. We also present a modular baseline solution for this task. One limitation of our work is the assumption of an existing human heatmap. Additionally, the lack of significant auditory background noise in both our simulation and real-world environments is notable as significant background noise could either prevent an emergency response (i.e. drowning out the trigger noise) or cause many extraneous explorations (i.e. construction noises making thud/thunk noises). As with many previous AudioGoal works, we leave explorations on the effects of background noise for future work. Privacy concerns connected to developing and storing a heatmap of user activity in the home, alongside the general collection of audio and visual data, would need to be considered and mitigated.

\textbf{Future Work.} While falls and fires cover many cases of home injuries or deaths, there are many more classes which could be added in the future to further address potential deaths. Understanding the scene completely and determining the source of the audio instead of screening for emergencies would be a valuable extension.

\textbf{Acknowledgments.} The authors want to thank Megha Maheshwari and Miguel Clement for their advice and technical assistance during the completion of this work.